\documentclass[twoside,twocolumn,9pt]{article}

\usepackage{blindtext} 

\usepackage{mathptmx}
\usepackage[T1]{fontenc} 
\usepackage{microtype} 

\usepackage{graphicx} 

\usepackage[english]{babel} 

\usepackage[a4paper,left=0.71in,top=0.98in,right=0.71in,bottom=0.98in,columnsep=15pt]{geometry} 
\usepackage[hang, small,labelfont=bf,up,textfont=it,up]{caption} 
\usepackage{booktabs} 

\usepackage{enumitem} 
\setlist[itemize]{noitemsep} 

\usepackage[runin]{abstract} 
\abslabeldelim{:}

\usepackage{titlesec} 
\titleformat{\section}[block]{\large\bfseries}{\thesection.}{1em}{} 
\titleformat{\subsection}[block]{\large}{\thesubsection.}{1em}{} 
\titleformat{\subsubsection}[block]{\normalsize}{\thesubsubsection.}{1em}{} 

\usepackage{titling} 

\usepackage{authblk}

\usepackage{caption}
\DeclareCaptionLabelFormat{nospace}{#1#2}
\usepackage[utf8]{inputenc}   				 	
\usepackage{silence}  							

\usepackage{url}
\usepackage{letltxmacro}
\LetLtxMacro{\autocite}{\cite}
\LetLtxMacro{\textcite}{\cite}

\usepackage[fleqn]{amsmath}

\usepackage{xcolor}
\usepackage[capitalize]{cleveref}
\crefname{figure}{Figure}{Figures}

\usepackage{amsmath} 
\usepackage{amsthm}  
\usepackage{amsfonts}
\theoremstyle{plain}

\theoremstyle{remark}\newtheorem{remarkenv}{Remark}        
\newenvironment{remark}{\begin{remarkenv}}%
	{\hfill$\lozenge$\end{remarkenv}}            

\usepackage{booktabs}
\usepackage{tabularx}
\newcommand{\otoprule}{\midrule[\heavyrulewidth]}

\usepackage{xcolor}
\usepackage{tikz}
\usepackage{pifont} 
\definecolor{resultfail}{rgb}{1.,0,0}
\definecolor{resultacceptable}{rgb}{1,0.76.,0}
\definecolor{resultfair}{rgb}{0,0.69.,0.94}
\definecolor{resultgood}{rgb}{0,0.69.,0.31}
\newcommand{\mycircle}[2]{\tikz \path[draw=#1, fill=#2] (0, 0) circle (1ex);}
\newcommand{\mysquare}[2]{\tikz \path[draw=#1, fill=#2] (0, 0) rectangle (2ex, 2ex);}
\newcommand{\fail}{\mycircle{resultfail}{resultfail}}
\newcommand{\acceptable}{\mycircle{resultacceptable}{resultacceptable}}
\newcommand{\fair}{\mycircle{resultfair}{resultfair}}
\newcommand{\good}{\mycircle{resultgood}{resultgood}}
\newcommand{\pass}{\textcolor{green}{\ding{52}}}
\newcommand{\nopass}{\textcolor{red}{\ding{55}}}
\newcommand{\nonconformity}{\mysquare{black}{yellow}}
\newcommand{\observation}{\mysquare{black}{blue}}

\providecommand{\keywords}[1]{\textbf{KEY WORDS:} #1}

\pretitle{\begin{center}\huge\normalfont\MakeUppercase} 
\posttitle{\end{center}\vskip 1em} 
\title{Procedure for the Safety Assessment of an Autonomous Vehicle Using Real-World Scenarios} 

\author[1,2]{\large\bfseries Erwin de Gelder}
\author[1]{\large\bfseries Olaf Op den Camp}

\affil[1]{\normalsize\textit{TNO, Integrated Vehicle Safety, Helmond, The Netherlands (E-mail: erwin.degelder@tno.nl, olaf.opdencamp@tno.nl)}}
\affil[2]{\normalsize\textit{Delft University of Technology, Delft Center for Systems and Control, Delft, The Netherlands}}
\date{} 

\renewcommand{\maketitlehookd}{%
\begin{abstract}	
\noindent The development of Autonomous Vehicles (AVs) has made significant progress in the last years and it is expected that AVs will soon be introduced on our roads. An important aspect in the development of AVs is the assessment of their safety. As traditional methods for safety assessment of vehicles are not feasible to be performed for AVs within reasonable time and cost, new approaches need to be worked out. Among these, real-world scenario-based assessment is widely supported by many players in the automotive field. Scenario-based assessment allows for using virtual simulation tools in addition to physical tests, such as on a test track, proving ground, or public road.

We propose a procedure for real-world scenario-based road-approval assessment considering three stakeholders: the applicant, the assessor, and the (road or vehicle) authority. In this procedure, the applicant applies for the approval of one specific AV, the assessor assesses this AV and advises the authority, and the authority sets the requirements for the AV and decides whether this specific AV is approved for road testing. The challenges are as follows. Firstly, the tests need to be tailored to the operational design domain (ODD) and dynamic driving task (DDT) description of the AV. Secondly, it is assumed that the applicant does not want to disclose all of the detailed test results because of proprietary or confidential information contained in these results. Thirdly, due to the complex ODD and DDT, many test scenarios are required to obtain sufficient confidence in the assessment of the AV. Consequently, it is assumed that due to limited resources, it is infeasible for the assessor to conduct all (physical) tests. 

We propose a systematic approach for determining the tests that are based on the requirements set by the authority and the AV’s ODD and DDT description, such that the tests are tailored to the applicable ODD and DDT. Each test comes with metrics that enables the applicant to provide a performance rating of the AV for each of the tests. By only providing a performance rating for each test, the applicant does not need to disclose the details of the test results. In our proposed procedure, the assessor only conducts a limited number of tests. The main purpose of these tests is to verify the fidelity of the results provided by the applicant. Still, the applicant needs to conduct many tests, but this is assumed to be feasible because the applicant can utilize virtual simulation tools to conduct (a part of) the tests whereas this is not possible for the assessor due to proprietary or confidential information. To illustrate the proposed procedure, an example is presented.

\keywords{safety assessment, autonomous vehicle, scenario-based assessment}
\end{abstract}
}

\begin{document}
	
\maketitle

\section{Introduction}
\label{sec:introduction}

The development of Autonomous Vehicles (AVs) has made significant progress. It is expected that AVs will be mainstream by 2040 \cite{madni2018autonomous} or earlier \cite{bimbraw2015autonomous}. 
An important aspect in the development of AVs is the safety assessment \cite{bengler2014threedecades, stellet2015taxonomy, Helmer2017safety, putz2017pegasus, wachenfeld2016release}. For legal and public acceptance of AVs, a clear definition of system performance is important, as are quantitative measures for the system quality. The more traditional methods \cite{ISO26262, response2006code}, used for evaluation of driver assistance systems, are no longer sufficient for the assessment of the safety of higher level AVs, as it is not feasible to complete the quantity of testing required by these methodologies \cite{wachenfeld2016release}. Therefore, the development of assessment methods is important to not delay the deployment of AVs \cite{bengler2014threedecades}.

One of the many challenges regarding the assessment of an AV \cite{koopman2016challenges} is to agree on a procedure that results in a reliable evaluation of the AV, provided that:
\begin{itemize}
	\item the assessment is sufficiently tailored to the Operational Design Domain (ODD) and Dynamic Driving Task (DDT) of the AV,
	\item the proprietary and confidential information regarding the development of the AV are respected, and
	\item the resources are limited. 
\end{itemize}
In this paper, we propose a procedure for a assessment of an AV that takes into account the aforementioned considerations. The procedure assumes a scenario-based approach for assessing the safety \cite{putz2017pegasus, winner2017pegasus, nhtsa2018framework}. In our procedure, three stakeholders are considered: the applicant that provides the AV, the authority that decides on the approval of the AV for road testing, and the assessor that performs the independent safety assessment of the AV. Based on the requirements set by the authority, the applicant and the assessor need to come to an agreement on the set of tests for the safety assessment. Based on the test results from both the applicant and the assessor, the assessor evaluates whether the AV is ready for deployment on the road and, if so, under which conditions.

To be best of the authors' knowledge, there is no literature that provides a procedure for the assessment of AVs while distinguishing between the different stakeholders that are involved. The proposed procedure could be used within a legal framework for the approval of AVs for road testing. 
Eventually, as technology improves, the procedure might be a good starting point for developing the legal framework for the type-approval of AVs.
In case of consumer testing performed by a New Car Assessment Programme (NCAP), the result is not an approval, but the proposed procedure could be used as well. 

In \cref{sec:problem}, we provide the problem definition after elaborating more on the context. The proposed procedure is presented in \cref{sec:procedure}. Using an hypothetical example in \cref{sec:example}, the procedure is illustrated. We end this work with conclusions in \cref{sec:conclusions}.

\section{Problem definition}
\label{sec:problem}

In this section, we first explain why many players in the automotive field support scenario-based testing for the assessment of performance aspects of the automated and autonomous vehicles, such as the safety assessment. Next, in \cref{sec:stakeholders}, we elaborate on the different (type of) stakeholders that are involved in the assessment. Given an assessment with these stakeholders, in \cref{sec:challenges}, we describe the problem and the corresponding practical challenges that our procedure should address.

\subsection{Scenario-based testing}
\label{sec:scenario-based testing}

Perhaps the most basic way of assessing the performance of an AV is to drive with the AV in its intended area of operation. While this might provide useful data for further development of the AV, \textcite{kalra2016driving,wachenfeld2016release} show that the number of hours of driving that are required to demonstrate with enough certainty that the AV performs safely is infeasible. So, when it comes to demonstrating the reliability of an AV, another approach is necessary.

An advantage of scenario-based testing is that it allows for selecting those scenarios that are relevant for the safety evaluation. Therefore, a large repetition of scenarios that are relatively straightforward to deal with can be prevented. Furthermore, because virtual simulations can potentially be used for performing scenario-based tests, the number of physical tests can be reduced \autocite{ploeg2018cetran}, ultimately resulting in a less expensive assessment. However, one of the main challenges of scenario-based testing is the selection of the scenarios itself \autocite{riedmaier2020survey}. 

Although there are challenges to be resolved for scenario-based testing, it is used in large research projects and by many players in the automotive field. For example, in European projects such as AdaptIVe \autocite{roesener2017comprehensive}, ENABLE-S3 \autocite{leitner2019validation}, and HEADSTART \autocite{wagener2020headstart}, scenario-based testing is proposed for assessing several performance aspects, such as safety and emissions. In Germany, a large project named PEGASUS was fully dedicated to scenario-based assessment of automated driving functions \autocite{pegasus2019}. Also, in Japan \autocite{jacobo2019development} and in Singapore \autocite{cetran2020}, a scenario-based approach is adopted for the assessment of AVs. To support the scenario-based assessment, different initiatives are started to create a database of scenarios and test cases, see, e.g., \autocite{elrofai2018scenario,myers2020pass}.

Considering the adoption of and the many resources dedicated to scenario-based assessment, it seems convincing that scenario-based assessment is a promising method for a framework for the safety assessment of AVs.

\subsection{Stakeholders for the assessment}
\label{sec:stakeholders}

We consider three different types of stakeholder for the assessment of an AV for its readiness to be deployed in a certain operational area:
\begin{enumerate}
	\item The applicant;
	\item The assessor; and
	\item The authority.
\end{enumerate}

The applicant is applying for the approval for the deployment of an AV for testing on the road. In practice, the applicant can be, e.g., the operator of the vehicles or the developer of the vehicle. The authority is the stakeholder that eventually decides whether the AV can be deployed, so this can be the local vehicle authority. It is a prerequisite for a proper process that the assessor is independent of both the applicant and the authority.
Each stakeholder has different responsibilities, see \cref{tab:stakeholders}. The applicant provides the AV to be assessed. The applicant is also responsible for providing an AV that meets the applicable safety requirements. The assessor is responsible for performing the tests in the assessment, not for deciding whether the AV is approved or not. Instead, the assessor advises the authority whether the AV can be approved and, if necessary, under which conditions. It is then up to the authority to make the final decision. Another responsibility of the authority is to organize a legal framework for the safety assessment of AVs and to set and communicate the requirements for the AV.

\begin{table}
	\centering
	\caption{The stakeholders and their responsibilities.}
	\label{tab:stakeholders}
	\begin{tabularx}{\linewidth}{lX}
		\toprule
		Stakeholder & Responsibilities \\ \otoprule
		Applicant & Apply for approval to deploy the AV. \\
		& Prepare the AV to meet the applicable safety requirements. \\
		& Provide the AV to be assessed. \\ \otoprule
		Assessor & Perform an independent safety assessment of the AV. \\
		& Report results and advise the authority for AV approval. \\ \otoprule
		Authority & Organize a legal framework for safety assessment of AVs. \\
		& Specify needs and set realistic AV safety requirements. \\
		& Decide on the approval of the AV. \\
		\bottomrule
	\end{tabularx}
\end{table}

\begin{remark}
	Within the terminology of the United Nations Economic Commission for Europe (UNECE), the assessor is called ``technical service''.
\end{remark}

\begin{remark}
	The presented stakeholders are particularly applicable for non-US approach. Currently, the role of the applicant and the assessor is often represented by one stakeholder in the US. Nevertheless, the presented procedure might still be applicable if a distinction is made between these two roles within the organization of the applicable stakeholder.
\end{remark}

\subsection{Challenges}
\label{sec:challenges}

Scenario-based assessment comes with many challenges. For example, questions like ``How to generate the test cases?'' and ``How to validate the fidelity of the virtual simulations?'' are discussed extensively in literature. In this paper, however, we want to specifically address the challenges that arise when considering the different stakeholders and each of their responsibilities and capabilities. Therefore, the problem we address in this paper can be formulated as follows:

\emph{What procedure could be used for the safety assessment of an Autonomous Vehicle (AV) by an independent assessor?}

In the framework presented in this paper, the following challenges are addressed:
\begin{itemize}
	\item The tests need to be tailored to the operational design domain (ODD) and dynamic driving task (DDT) description of the AV. However, it is expected to be infeasible for the assessor to go through a rigorous analysis to define all relevant tests for each applicant.
	\item It is assumed that the applicant does not want to disclose details of sensor and system implementation or even detailed test results because of proprietary or confidential information contained in these results. As a result, the assessor does not have access to these detailed test results. The challenge is that the assessor still needs enough confidence in the safe operation of the AV without having access to the detailed test results carried out by the applicant.
	\item Due to the complex ODD and DDT, it is expected that many tests are required to obtain enough confidence in the assessment of the AV. Also, it is assumed that the assessor's resources are too limited to conduct all tests physically. The challenge is that the procedure should still allow the assessor to have enough evidence that the AV is safe or not.
\end{itemize}

\section{Procedure for the safety assessment}
\label{sec:procedure}

\begin{figure*}
	\centering
	\includegraphics[width=\linewidth]{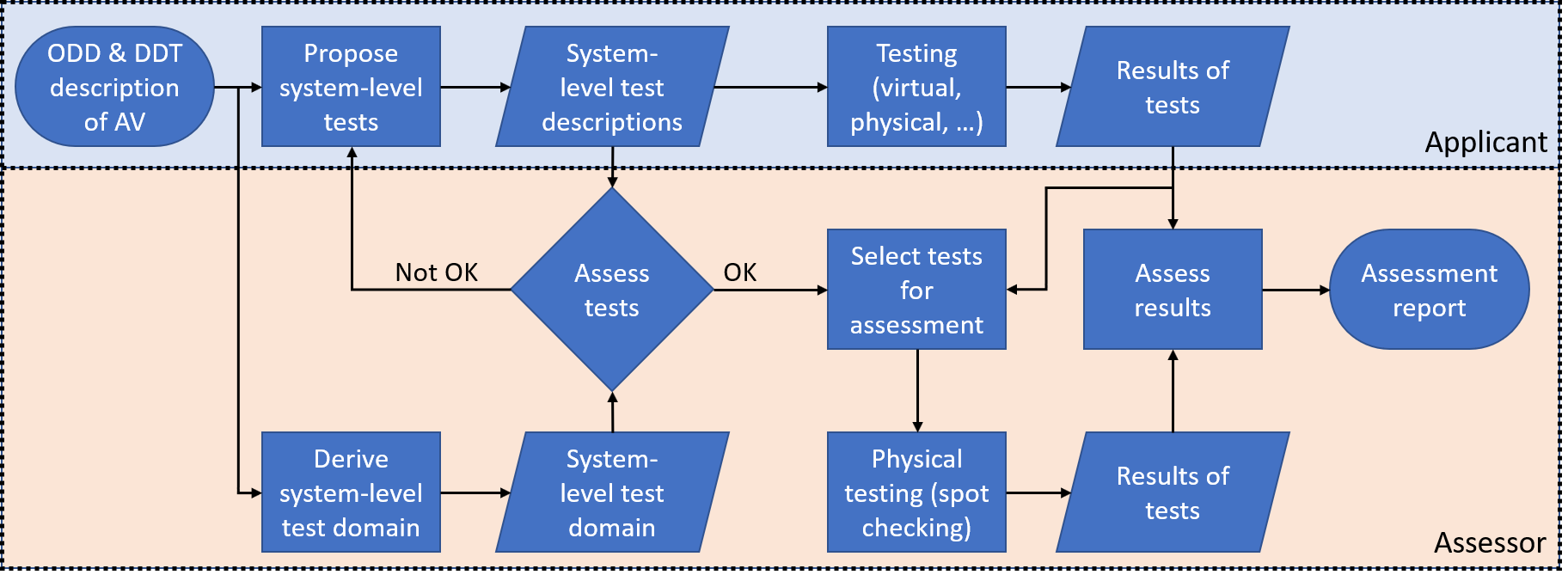}
	\caption{Proposed framework for the safety assessment of an autonomous vehicle (AV).}
	\label{fig:procedure}
\end{figure*}

This paper assumes that many of the relevant tests for the safety assessment are performed in a virtual simulation environment that is controlled by the applicant. The proposed procedure intends to consider all results, both from virtual simulation and from actually performed physical tests. Where the assessor does not have access to the required models of the AV under test, the assessor will have the capability to perform physical tests on the AV. How to balance between the different results in the assessment, considering virtual and physical test results of the applicant and physical test results of the assessor is schematically presented in \cref{fig:procedure}. Each rectangular block represents an action. The procedure distinguishes between actions for which the applicant is responsible and actions for which the assessor is responsible. The procedure consists of the following actions:
\begin{enumerate}
	\item The first action is to derive which system-level tests need to be performed with reference to the ODD and DDT of the AV under test. Here, “system-level” is mentioned explicitly, because it is assumed that also in case of a failure of any of the subsystems, the AV would fail the system-level tests. Note, however, that it is advised that the applicant ensures that each of the subsystems underwent a rigorous assessment before applying for the AV assessment. 
	\item If the derived tests are acceptable, the next action is to select the tests for the assessment. Here, a distinction is made between tests for which the applicant is fully responsible and physical tests that are conducted by the assessor. The latter will focus more on spot checking.
	\item Once the tests are selected, these tests need to be conducted. The results of these tests will be described using prescribed metrics. Note, however, that these metrics may not contain too much information as it is assumed that the applicant does not want to disclose details of sensor and system implementation or even detailed test results because of the proprietary or confidential information contained in these results. 
	\item The final step is to assess the results from the tests and to formulate an advice for the authority on whether the AV is ready for deployment and under which conditions.
\end{enumerate}

In the following sections, each of the actions are further detailed. We end this section with a short note on monitored deployment in case of a successful completion of the assessment.

\subsection{Deriving test descriptions}
\label{sec:test descriptions}

Based on the ODD and the DDT of the AV, the tests are derived. Following the same reasoning as \textcite{stellet2015taxonomy}, a test is an evaluation of:
\begin{itemize}
	\item a statement on the system-under-test (test criteria; what are we going to evaluate using the test);
	\item under a set of specified conditions (test case; how are we going to evaluate the test criteria);
	\item using quantitative measures (metrics; how to express the outcome of the test quantitatively);
	\item and a reference of what would be the acceptable outcome (reference; when is the outcome acceptable).
\end{itemize}

Since the applicant has designed and developed the AV, it is expected that the applicant has a clear notion of the tests that are required for a complete assessment of the AV and which the AV should appropriately handle. Similarly, if the set of relevant test descriptions is not complete during the development of the AV, it is conceivable that the AV will not operate safely for all circumstances possible within the ODD. 

Although it is expected that the applicant provides all relevant test descriptions, it is important that the applicant and the assessor discuss these test descriptions, and that a check is made whether or not the test descriptions are complete and cover the ODD sufficiently. If an important test description is missing, it is conceivable that the AV is not specifically designed to pass the corresponding test. In order to assess the completeness of the test descriptions provided by the applicant, the assessor needs to define the test domain for the relevant system-level test descriptions and use these to investigate if any important test descriptions are missing. Here, the so-called test domain refers to a more high-level description of the range of tests that are expected, rather than an enumeration of the large number of relevant tests.

If it turns out that the test descriptions that are provided by the applicant are not complete, the process needs to be restarted, as indicated by the ``Not OK'' line in \cref{fig:procedure}. On the other hand, if the test descriptions are deemed to be complete enough, the assessment proceeds to the next step: selecting tests for the assessment.

\subsection{Selecting tests for the assessment}
\label{sec:selection}

In principle, the applicant is expected to provide results for all tests. In the next section, we explain how these results may look like. Based on these results, tests are selected for the physical testing performed by the assessor. This is indicated by the arrow pointing from ``results of tests'' of the applicant to ``select tests for assessment'' in \cref{fig:procedure}.  

\begin{figure}
	\centering
	\includegraphics[width=.65\linewidth]{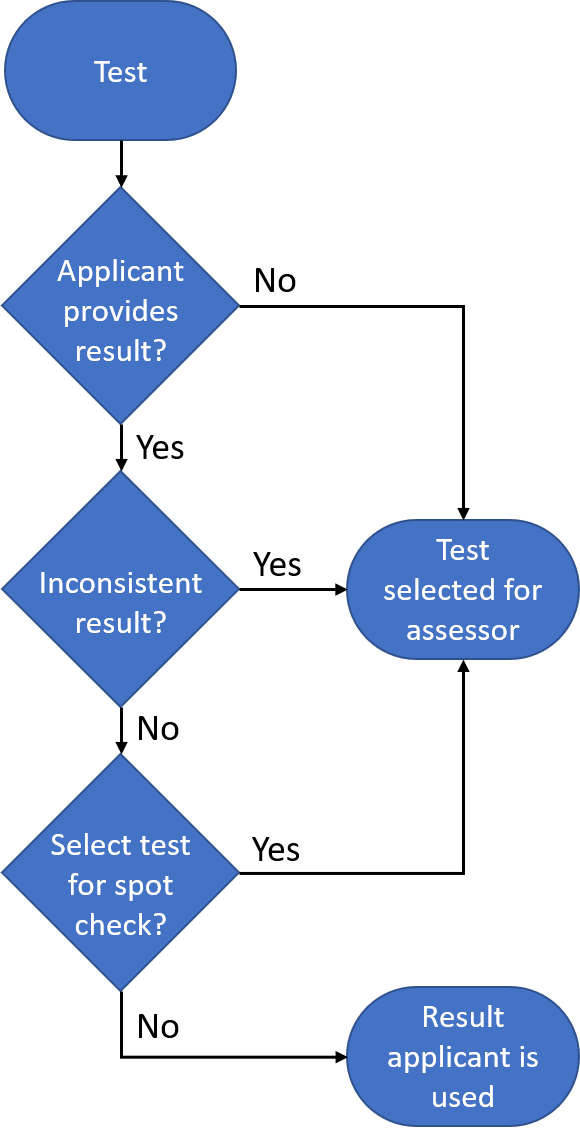}
	\caption{Decision scheme for the selection of a test for physical assessment by the assessor.}
	\label{fig:selection}
\end{figure}

A test is selected for physical testing by the assessor if any of the following three statements are true:
\begin{itemize}
	\item The applicant does not provide a result. Although the applicant is expected to provide results for most tests, it might be possible that there are some tests for which the applicant does not have the resources to perform the tests reliably, for example if specific tooling is required. Note, however, that if the applicant does not provide results for too many tests, the assessment automatically results in a fail.
	\item The result seems inconsistent. If there is sufficient reason for the assessor to not confide in the result provided by the applicant, the test can be performed by the assessor to check the result provided by the applicant.
	\item The test is selected for spot checking. The main reason to perform spot checking is to assess the fidelity of the results provided by the assessor.
\end{itemize}
The process of test selection is summarized in \cref{fig:selection}.

\subsection{Testing}
\label{sec:testing}

\begin{figure}
	\centering
	\includegraphics[width=\linewidth]{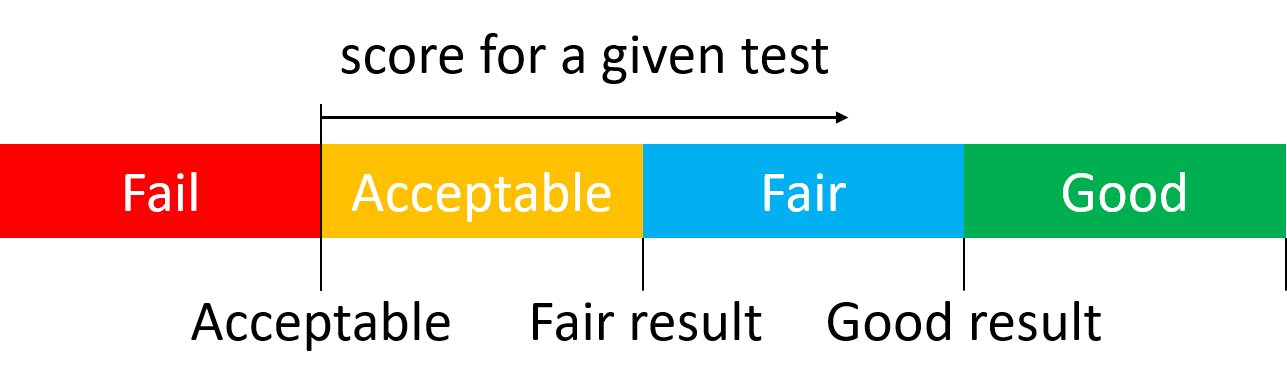}
	\caption{Different scoring options per test.}
	\label{fig:rating}
\end{figure}

As explained in the previous section, the applicant is expected to provide results for most tests. However, it is assumed that the applicant does not want to disclose detailed test results. Therefore, a rating scheme is proposed. Using a specific metric for each test, three references are defined: an acceptable result, a fair result, and a good result. If the result of the test is worse that the acceptable result, a ``fail'' is reported. If the result passes the acceptable reference but not the what is defined as a fair result, an ``acceptable'' is reported. Similarly, a ``fair'' is reported if the result is between a fair and a good result. If the result is better than what has been defined as a good result, a ``good'' is reported. This is schematically shown in \cref{fig:rating}. 

In principle, the applicant is free to choose any method to derive the results. However, considering the large number of tests, the use of virtual simulations seems inevitable. In practice, it is expected that a both virtual simulations, physical tests, and a combination, such as hardware-in-the-loop testing, is used to determine the test results.

On the other hand, the tests by the assessor are performed physically. The main reason for this is that virtual simulations are ruled out as that would require the applicant to provide a model of the AV, which is expected to be impossible because of proprietary reasons.

\subsection{Assess results}
\label{sec:assess results}

The following assessment results are distinguished per test:
\begin{itemize}
	\item In case the test results show that for the specific test the AV performs acceptable (i.e., ``acceptable'', ``fair'', or ``good'', see \cref{fig:rating}), the test is passed. If this is not the case, then the specific test fails.
	\item Inspired by IATF~16949 on automotive quality management \autocite{IATF16949}, a passed test may result in a non-conformity. An “acceptable” result automatically leads to a non-conformity (NC). This means that the response of the AV deviates substantially from  response that is qualified as “good”, but the deviation is not severe. Since the AV meets the minimum requirement for this test and consequently safety is not compromised, there is no reason to fail the AV based on this test. Nevertheless, an NC is issued to indicate that the applicant is asked to consider improvements, e.g., for a next version of the system.
	\item In case the test is also performed by the assessor and the corresponding result is worse than the reported result of the applicant, this also leads to a NC.
	\item The assessment of a test result might come with an observation (OB) that needs consideration of the applicant. 
\end{itemize}

If a test results in a fail, then either the assessment results in a negative advice of the assessor to the authority or it is advised to only allow for deployment of the AV under certain conditions. For example, if the only tests that are failed consider low-light conditions, the AV might be deployed under the condition that it operates only from sunrise till sunset. 

NCs and OBs do not lead to an immediate fail of the assessment. However, it is likely that they lead to a fail in a future assessment, e.g., when test criteria become increasingly demanding, and the applicant does not appropriately consider such NCs or OBs. NCs provide information to the applicant on how requirements might develop in the future, which, consequently, gives direction and motivation on continuous improvement of AVs regarding safety. On the other hand, many NCs – the AV barely passes the test in many cases – might mean that safety is compromised and, therefore, it might also result in a negative advice of the assessor to the authority regarding the deployment of the AV.

Note that when many NCs are observed, the AV probably will not be able to pass all tests if all tests would be performed physically by the assessor. Theoretically, this is however still possible. To minimize the risk of having an AV that passes all tests, but with many NCs, a system using demerit points is introduced. The AV starts with, e.g., 100 points, and in the assessment, 1, 2, or 3 points are subtracted for each NC, depending on the severity of the NC. Once the number of points for the AV are reduced to 0, then the AV is indicated to have failed he assessment because of an overrun of NCs. The numbers given here are merely provided as an example.

\subsection{Monitored deployment}
\label{sec:monitored deployment}

A successful completion of the proposed assessment might result in the approval for the deployment of the AV under the condition that the behavior of the AV on the road is continuously monitored. We propose that during such a deployment phase, the applicant is required to upload detailed driving data to allow for monitoring the AV behavior. This is implemented for two reasons:
\begin{itemize}
	\item After completion of the assessment pipeline, road and/or vehicle authorities may require the monitoring of safety continuously when driving on the public road.
	\item The uploaded data may be used to improve the generation of tests and the selection of relevant test cases for a particular AV, as is possible that some tests have been overlooked during the initial assessment process or that situations on the road gradually change with changes in traffic, e.g.because of the introduction of new mobility systems.
\end{itemize}

The feedback to the data acquisition element allows for ongoing learning and improvement of the standards and assessment systems, while being able to adapt to new types of transportation such as personal mobility devices. For example, additional test cases could be identified and incorporated into future safety assessment procedures. A deployment might consider new operational areas, the extension of the scenario database with scenarios that potentially differ between such areas would then be covered. Moreover, to obtain a scenario database that is `complete', i.e., statistically accurate, it is expected that operational data collection is required over an extended period, which most probably will not be realized before the deployment is operationalized. In other words, the imperfection of the assessment framework should not become a barrier for the introduction of new safe mobility solutions onto the market, in case these devices are tested to be safe for all currently known conditions. The assessment method, especially the step regarding monitored deployment, supports the continuous increase in knowledge on the state-of-the-art of road safety and herewith prepares the safety assessment method to be sustainable for the future.

\section{Example}
\label{sec:example}

In this section, we present a hypothetical example to illustrate how the safety assessment as discussed in the previous section may look like. \Cref{tab:example} lists the results of an assessment consisting of 14 tests. Note that in reality, the number of tests is likely to be much larger, but for the sake of the example, we keep the number of tests rather limited. 

\begin{table}
	\centering
	\caption{Results from a hypothetical assessment. The result may be ``fail'' \protect\fail, ``acceptable'' \protect\acceptable, ``fair'' \protect\fair, or ``good'' \protect\good; \protect\pass{} denotes a pass and \protect\nopass{} denotes a fail; NC and OB are denoted by \protect\nonconformity{} and \protect\observation{}, respectively.}
	\label{tab:example}
	\begin{tabular}{llllll}
		\toprule
		Test            & Result      & Result      & Fidelity & P/F     & NC  \\
		ID              & applicant   & assessor    & check    &         & OB  \\
		\otoprule
		1.1             & \good       & \good       & \pass    & \pass   &     \\
		1.2             & \fair       & -           & \pass    & \pass   &       \\
		1.3             & \fair       & -           & \pass    & \pass   &        \\
		1.4             & \acceptable & -           & \pass    & \pass   &\nonconformity    \\
		2.1             & \good       & -           & \pass    & \pass   &        \\
		2.2             & \good       & \fair       & \nopass  & \pass   & \nonconformity      \\
		2.3             & \fair       & -           & \pass    & \pass   &        \\
		2.4             & \fair       & \fair       & \pass    & \pass   &     \observation \\
		3.1             & \fair       & \acceptable & \nopass  & \pass   & \nonconformity     \\
		3.2             & \fair       & \good       & \pass    & \pass   &       \\
		3.3             & \fail       & -           & \pass    & \nopass &       \\
		4.1             & \fair       & -           & \pass    & \pass   &       \\
		4.2             & \acceptable & -           & \pass    & \pass   & \nonconformity      \\
		4.3             & \acceptable & \fail       & \nopass  & \nopass &       \\ \bottomrule
	\end{tabular}
\end{table}

The applicant reports a good result for the first test (test ID 1.1). Because the assessor comes to the same conclusion, the fidelity check is passed as well as the test. The next three tests (1.2 till 1.4) are not performed by the assessor, so the reported results of the applicant are included in the assessment result. These three tests are all passed, but because the last of these tests (1.4) barely passes the minimum requirement (an “acceptable” is reported), an NC is issued.

The tests 2.1 till 2.4 are all passed. However, the applicant reports a better result for test 2.2 than the assessor. Therefore, the fidelity check failed, and an NC is issued. For test 2.4, an OB is made. This could be, e.g., because during the test that is performed by the assessor the AV showed erratic behavior even though safety has not been compromised.

For the next three tests (3.1 till 3.3), an NC is issued, and one test is failed. The NC is issued for two reasons: the assessor reports an acceptable result and the fidelity check has failed. The applicant reports a fail for test 3.3. There is no reason for the assessor to also perform this test, because regardless of that result, the applicant has failed the test. As mentioned in Section 3.4, this might lead to a failed assessment or to some restrictions for the deployment of the AV. Note that for test 3.2, the result of the assessor is better than the reported result of the applicant. This might be caused by the applicant reporting the worst-case outcome, e.g., the outcome with a relatively long detection delay of an object while the detection delay is shorter during the test performed by the assessor. Because the result of the assessor is better, this does not lead to an NC.

For the last three tests, again an NC is issued, and a test is failed. The NC is issued because the applicant reports an acceptable result. Test 4.3 is failed even though the applicant has not reported to fail this test. This might lead to additional measures that need to be taken before the applicant can proceed with the deployment of the AV.

\section{Conclusions}
\label{sec:conclusions}

We proposed a procedure for the assessment of an Autonomous Vehicle (AV) in order to answer the question ``What procedure could be used for the safety assessment of an AV by an independent assessor?''
In the proposed procedure, a distinction is made between activities of the applicant and the independent assessor, while considering the limited resources and the proprietary and confidential information related to the development of the AV.

There are still some open questions. E.g., how to assure that the set of tests cover the ODD and DDT sufficiently? Also, since virtual simulations seem inevitable, how to prove that the results of the virtual simulations alongside the limited number of physical tests provide a reliable evaluation of the AV?

Despite these open questions, the proposed procedure can be used as a starting point for discussions on the development of a legal framework for authorities for the safety assurance of AVs before deploying these vehicle on the public road.


\bibliographystyle{ieeetran}
\bibliography{bib}


\end{document}